\documentclass[10pt,journal,compsoc]{IEEEtran}
\usepackage{blindtext}
\usepackage{amsmath}
\usepackage{graphicx}
\usepackage{biblatex}
\usepackage{bbm}
\usepackage{algorithm2e}  
\usepackage{setspace}
\usepackage{algorithmic}

\DeclareMathOperator{\supp}{supp}

\begin{document}
%
\title{Task Specific Adversarial Cost Function}

\author{Antonia~Creswell,~\IEEEmembership{Student Member,~IEEE,}
        and Anil~A.~Bharath,~\IEEEmembership{Member,~IEEE}
\IEEEcompsocitemizethanks{\IEEEcompsocthanksitem{A. Creswell and A.A. Bharath are with Imperial College London.}}
\thanks{A. Creswell wishes to thank...}}

\markboth{Submitted to IEEE T-PAMI}%
{Creswell: Task Specific Cost Function}

\IEEEtitleabstractindextext{%

\begin{abstract}
The cost function used to train a generative model should fit the purpose of the model. If the model is intended for tasks such as generating perceptually correct samples, it is beneficial to maximise the likelihood of a sample drawn from the model, $Q$, coming from the same distribution as the training data, $P$. This is equivalent to minimising the Kullback-Leibler ($KL$) distance, $KL[Q\|P]$. However, if the model is intended for tasks such as retrieval or classification it is beneficial to maximise the likelihood that a sample drawn from the training data is captured by the model, equivalent to minimising $KL[P\|Q]$. The cost function used in adversarial training optimises the Jensen-Shannon entropy which can be seen as an even interpolation between $KL[Q\|P]$ and $KL[P\|Q]$. Here, we propose an alternative adversarial cost function which allows easy tuning of the model for either task. Our task specific cost function is evaluated on a dataset of hand-written characters in the following tasks: Generation, retrieval and one-shot learning.






\end{abstract}

}



\maketitle


%
\IEEEpeerreviewmaketitle

\IEEEraisesectionheading{\section{Introduction}\label{sec:introduction}}
\subsection{Generative and Discriminative Models}

\IEEEPARstart{D}{iscriminative}  models are trained to predict a label, $y$, given an input sample, $x$. Probabilistically, this is equivalent to learning a conditional probability $p(y|x)$. State-of-the-art discriminative models are able to outperform humans on tasks such as natural image recognition \cite{szegedy2015going} and sketch recognition \cite{yu2015sketch}. However training models to achieve or exceed human levels of recognition requires large amounts of labelled training data, which if often expensive to acquire. 

Recently, there has been immense interest in generative models, which are able to learn form unlabelled training data, which is often available in abundance.  However, generative models are more challenging to learn. A generative model should be able to draw samples from $p(x)$; however estimating $p(x)$ may be computationally intractable. Instead, we often learn a function that maps a vector to an image sample $x$. The vector may be either be a noise vector, $z$, drawn from a prior distribution \cite{kingma2013auto,radford2015unsupervised}, a label vector \cite{dosovitskiy2015learning}, $y$, or a combination of the two \cite{chen2016infogan, mirza2014conditional,gauthier2014conditional, vincent2008extracting}. Probabilistically, these may be interpreted as conditional probabilities: $p(x|z)$, $p(x|y)$  or $p(x|z,y)$. By sampling these conditional probabilities appropriately, novel samples of $x$ may be generated.

Generative models are not only useful for synthesising new samples, they may also learn a representation for the training data that can be applied to discriminative tasks, via semi-supervised learning \cite{lake2015human,chen2016infogan, radford2015unsupervised}. Semi-supervised learning makes use of large amounts of accessible, unlabelled data to train a model that learns a representation for the data. A smaller set of labelled samples may be mapped to the learned representation space, which hopefully makes classes more separable, allowing a discriminative classifier to be trained using few labelled samples. 

There is currently active research in applying generative models to image data, both to improve the quality of generated images \cite{oord2016pixel,zhao2016energy,chen2016infogan,dosovitskiy2015learning} and to apply representations learned during training to discriminative, image tasks \cite{radford2015unsupervised,lake2015human}.

\subsection{Image Synthesis Using Generative Models}
Auto-encoders learn an encoder, $p(h|x)$, which maps from image space to a latent space and a decoder, $p(x|h)$ which maps back to image space. Auto-encoders are trained to reconstruct samples, rather than synthesise new samples; this is because the distribution, $p(h)$, of the latent space is unknown, and so the decoder cannot be sampled. 

Variational auto-encoders \cite{kingma2013auto} address this problem by constraining $h$ to come from a prior distribution e.g. a normal distribution. Variational auto-encoders can be implemented by first sampling from the prior and then sampling the conditional probability $p(x|h)$ to get a new sample. Samples generated using variational auto-encoders are often overly smoothed because of the constraint on the latent space. But it is not always necessary to constrain the latent space of an auto-encoder in order to generate new samples. 

De-noising auto-encoders \cite{bengio2013generalized} are trained to reconstruct an image from a corrupted version. If the corrupted image, $x'$ is sampled from the conditional probability $p(x'|x)$ and the de-noising auto-encoder samples $p(x|x')$, new samples may be generated by alternatively sampling $x'_{t} \sim p(x'|x)$ and $x_{t+1}\sim p(x|x'_{t})$, where a new sample is generated at each time step, $t$.

The generative models described thus far look at generating an entire image in one go. An alternative is to develop a single image sequentially. Gregor et al. \cite{gregor2015draw} learn to generate hand-written digits sequentially using an auto-encoder architecture with recurrent connections and an attention mechanism. The attention mechanism allows the generator to focus on smaller regions of an input image, and generate an image a few pixels at a time. The sequential approach generates sharper samples than traditional variational auto-encoders \cite{kingma2013auto}. Gregor et al. also modified their approach to generate two digits per image, and observed that their attention mechanism ensured that the model focused on generating one number at a time. 

A more extreme approach to sequential generation is to generate images pixel by pixel. Oord et al. \cite{oord2016pixel} generated natural images one pixel at a time, where the choice for the next pixel depended on the previous pixels. Though the ``natural" images generated by Oord et al. \cite{oord2016pixel} do not resemble class specific samples, the statistics of ensembles of samples appear to be consistent with those of natural images.

Lake et al. \cite{lake2015human} also generated images sequentially, by using labelled data within a Bayesian probabilistic method for generating hand-written characters, one stroke at a time. Lake et al. \cite{lake2015human} were able to synthesize very sharp image samples that closely resembled real image examples.

An approach to learning generative models by using labels was suggested by Dosovitskiy et al. \cite{dosovitskiy2015learning}. The authors trained a convolutional neural network to generate images of tables, cars and chairs from a series of vectors that encoded object class, viewpoint and a spatial transform. They were able to generate examples of objects from varying viewpoints and morph different styles of chairs to suggest new chair designs.

\subsection{Discriminative Tasks On Images Using Generative Models}
De-noising auto-encoders \cite{vincent2008extracting} can be trained on unlabelled data to learn a representation. Training involves learning both an encoder and a decoder; once trained, the decoder may be removed and the network architecture modified for classification. The network may then be \textit{fine-tuned} by training the network for the classification task. It is suggested by Erhan et al. \cite{erhan2010does} that this process of \textit{pre-training} and fine-tuning prevents models that are designed for classification from over-fitting.

Lake et al. \cite{lake2015human} learned generative models for hand-written characters from labelled training data. However, they were able to apply their generative model to also find matching characters for queries from unseen classes. This type of learning -- from only one example -- is known as one-shot learning, and is a very challenging task. When shown a character from an unseen class, and presented with $20$ samples from other unseen classes (including a sample from a similar class), their model was able to pick the correct matching sample more accurately than humans \cite{lake2015human}.

A generative model that was first introduced by Goodfellow et al. \cite{goodfellow2014generative} and improved by Salimans et al. \cite{salimans2016improved} was able to achieve state-of-the-art recognition in semi-supervised classification on CIFAR-10 (a dataset of small natural images), MNIST (a dataset of hand-written digits) and SVHN (a dataset of street numbers).

\subsection{Task Specific Cost Function For Training Generative Models}
A generative model may learn to generate samples with distribution, $Q$, which captures the underlying probability distribution of the training data, $P$, by minimising a cost function that measure the difference between the two distributions.

\begin{figure}
    \centering
    \includegraphics[width=\columnwidth]{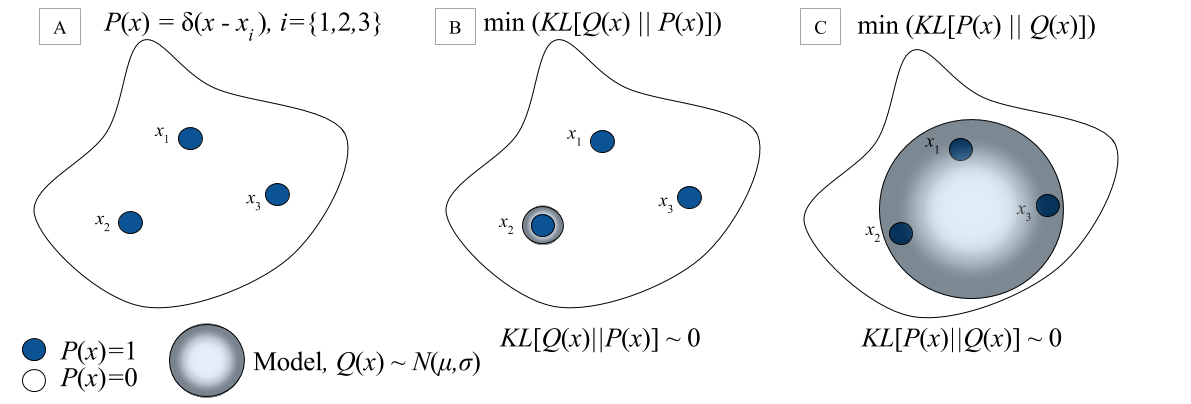}
    \caption{For a data distribution $P(x)=\sum_{i}{\delta(x - x_i)}$ for $i \in \{1,2,3\}$ a model, $Q(x)$ with finite capacity may be fit by minimising either A) $KL[Q\|P]$ which captures one region of high density well, but ignores others, or B) $KL[P\|Q]$ which captures all regions of high density while also assigning non-zero values to regions of low density.}
    \label{KL_dig}
\end{figure}

The cost function used to train a generative model should fit the purpose of the model. If the model is intended for generation of perceptually high quality samples, it is necessary for the model to capture the densest parts of the training data distribution. This can be achieved by learning a $Q$ that minimises $KL[Q\|P]$ \cite{huszar2015not}. For a model with sufficient complexity and many samples, it may be possible to learn a $Q$ such that $Q=P$. However, for a finite model, and with insufficient training samples, the model is likely to fit only the densest parts of the distribution at the cost of not capturing other regions of high density. A pictorial example of this is shown in Fig. \ref{KL_dig}A.

If the representation of a generative model is intended to be used for discriminative tasks such as retrieval or classification, it is necessary to learn a model that captures the whole distribution of the training data. To achieve this, $KL[P\|Q]$ may be minimised. A model, $Q$ with finite capacity will be penalised if it does not capture states in $P$, which will encourage the model to capture all regions of high density in the data distribution at the cost of also capturing regions of low density. The model $Q$ would be less suitable for generation because generations sampled from regions of low probability are likely to be nonsensical. A pictorial example of this is shown in Fig. \ref{KL_dig}B.

A cost function that may allow tuning towards one task or the other is the Jensen-Shannon divergence \cite{lin1991divergence}, $JS_\pi$:
\[JS_\pi = \pi KL[P\| \pi P + (1-\pi)Q] + (1-\pi) KL[Q \| \pi P + (1-\pi)Q] \] for $0<\pi<1$.

Huszar et al. \cite{huszar2015not} showed that $JS_\pi$ is proportional to $KL[P\|Q]$ and $KL[Q\|P]$, respectively, at the upper and lower limits of $\pi$:
\[\frac{JS_\pi[P\|Q]}{\pi} \rightarrow KL[P\|Q] \hspace{0.5cm} \textrm{as} \hspace{0.5cm} \pi \rightarrow 0\]
\[\frac{JS_\pi[P\|Q]}{(1-\pi)} \rightarrow KL[Q\|P] \hspace{0.5cm} \textrm{as} \hspace{0.5cm} \pi \rightarrow 1\]

In previous work Goodfellow et al. \cite{goodfellow2014generative} introduced adversarial training, where a pair of competing models -- a generator and discriminator -- are trained. The generator is trained to produce samples that appear to come from the training data, and the discriminator is trained to distinguish the training data samples from generated samples. Training is successful when the discriminator cannot distinguish synthesized samples from samples that are drawn from the training data. Goodfellow et al. \cite{goodfellow2014generative} applied adversarial training to learn a distribution over image space in order to synthesize new image samples. Further, Goodfellow et al. \cite{goodfellow2014generative} showed that under certain conditions adversarial training minimizes the Jensen-Shannon divergence at a fixed $\pi$ value of $\pi=0.5$; this is more commonly known at Jensen-Shannon entropy. By optimising the Jensen-Shannon entropy, $JS_{\pi=0.5}$ rather than the Jensen-Shannon divergence, $JS_{\pi}$, the training algorithm and cost function proposed by Goodfellow et al. \cite{goodfellow2014generative}, do not depend on $\pi$ and so cannot be tuned towards one task or the other.

The training algorithm proposed by Goodfellow et al. \cite{goodfellow2014generative} draws samples equally from $P$ and $Q$. Huzar \cite{huszar2015not} proposes an alternative training algorithm for approximating $JS_\pi$ for small and large $\pi$ values by using biased sampling, drawing more values from $P$ to approximate $KL[P\|Q]$ or more values from $Q$ to approximate $KL[Q\|P]$.  However, we are not aware of any experiments that explore the effects of such a sampling strategy, and only a qualitative relationship between the number of samples and the effect on the cost function was suggested. 


Instead of using biased sampling during training to approximate $JS_{\pi}$, we propose an alternative adversarial cost function which we show is equivalent to $JS_\pi$ plus additional terms that depend only on $\pi$. In the limits of $\pi$, the additional terms tend to zero and $JS_\pi$ tends towards $KL[P\|Q]$ or $KL[Q\|P]$, depending on the choice of $\pi$. The parameter $\pi$ may be chosen to suit the desired task. We apply our novel cost function to both discriminative and generative tasks to show that a smaller $\pi$ value improves performance on discriminative tasks while larger $\pi$ values improve performance on generative tasks.





\section{Preliminaries: Generative Adversarial Networks}
\begin{figure}
    \centering
    \includegraphics[width=\columnwidth]{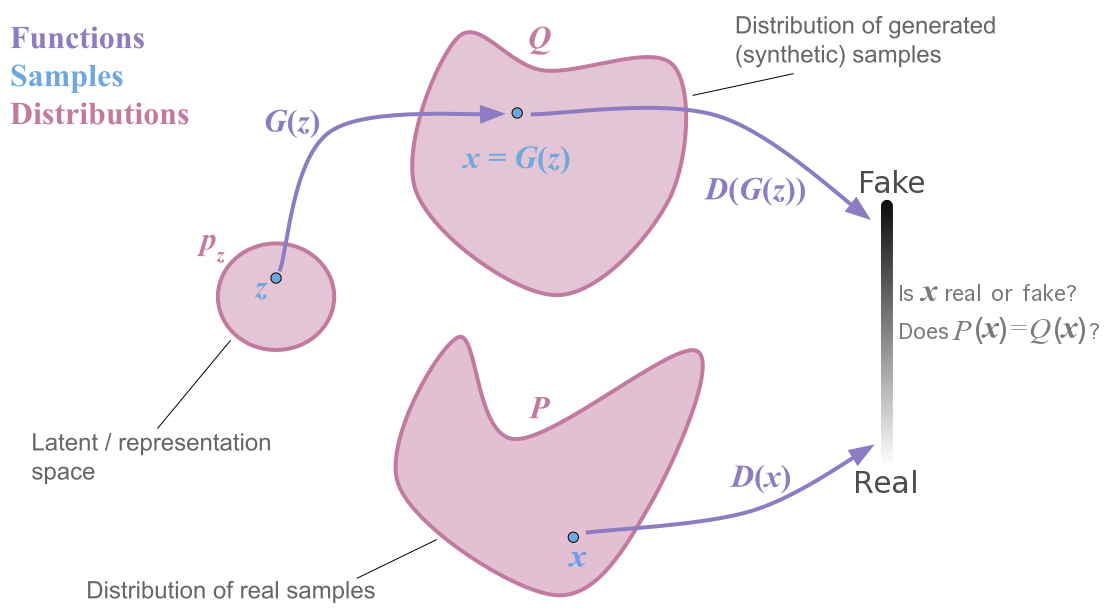}
    \caption{Generative Adversarial Networks in the context of generative models: A random sample $z$ is drawn from a prior distribution $p_z$ and mapped by $G$ to be a sample in the model distribution space, $Q$. Samples, $x$ from the training data distribution, $P$ or the model distribution $Q$ are mapped by $D$ to a $(0,1)$ prediction of whether the sample is from the training data distribution or not.}
    \label{gen_model}
\end{figure}

The purpose of image generation models is to learn a distribution $Q$ which captures a training data distribution $P$, over image space. Often, learning to draw samples from $Q$ directly is computationally intractable. Instead, we want to learn a parametrised function, $G(z; \theta_G)$ which maps samples, $z$, from a prior distribution $p_z(z)$ to an image sample in $Q$. During training, the parameters, $\theta_G$, are learned such that $Q$ is similar to $P$. This requires comparing samples $G(z;\theta_G)$ to real image samples. For example, in an auto-encoder \cite{larsen2015autoencoding}, this could be achieved by calculating the pixel-wise error between the generated and real samples (e.g. using MSE or cross-entropy). However, comparing pixel values to evaluate $G(x)$ has often been found to lead to poor quality image generation \cite{larsen2015autoencoding}. Instead, another parametrised function, $D(x, \theta_D)$, could be introduced to map all samples directly to a probability of whether that sample is likely to have come from the real data distribution or not, see Fig. \ref{gen_model}. This is the idea behind adversarial training, which we will now explain more formally.


In adversarial training a pair of networks is trained, a generator, $G$ and a discriminator, $D$. The generator takes as input a vector of random values, $z \in R^{n}$ drawn from a prior distribution $p_z(z)$. During training, one objective is to learn a mapping $G: R^{n} \rightarrow R^{M}$ from latent space to sample space, where $M$ is the dimensions of the sample space and $n$ is the scalar dimension of $z$. The discriminator takes samples either from the training dataset, $x \sim P$, or the generator, $x \sim Q$. During training, the discriminator is trained under a different objective: to learn a mapping $D: R^{M} \rightarrow (0,1)$, predicting a label for whether a sample was drawn from the training data, (1 - real) or from the generator, (0 - fake). The  objective function for training the discriminator is to correctly classify examples as being either real or fake. A well-trained generator can create samples that are realistic enough to fool the discriminator into making incorrect classifications. See Fig. \ref{gen_model}.

Previous work on adversarial training has primarily focused on either generating realistic looking samples, \cite{goodfellow2014generative, gauthier2014conditional, makhzani2015adversarial, kataokaimage}, classification tasks \cite{ganin2016domain, radford2015unsupervised, makhzani2015adversarial, chen2016infogan} or multi-label tagging \cite{mirza2014conditional}. More recently, adversarial training has also been applied to image retrieval \cite{creswell2016adversarial, donahue2016adversarial}. However, adversarial training optimises a cost function which approximates $JS_{\pi=0.5}$  divergence \cite{goodfellow2014generative}. The resulting generative model is not ideal for tasks such as classification or retrieval. We propose an alternative cost function that can be tuned to make adversarial training more suitable either for discriminative tasks or for generative tasks.


\section{Motivation (Previous Work)}

Generative Adversarial Networks (GANs) have recently attracted interest because of their ability to learn complex generative models with minimal labelling of data. Goodfellow et al. \cite{goodfellow2014generative} introduced GANs, modelling both the generator, $G$, and the discriminator, $D$, as fully connected neural networks. Radford et al. \cite{radford2015unsupervised} extended GANs by using fully convolutional neural networks for both $G$ and $D$. These convolutional networks are capable of generating images of realistic looking faces, bedrooms and numbers.

GANs may also be trained with labels \cite{mirza2014conditional, gauthier2014conditional}, such that images of specific categories may be generated. These networks are called conditional GANs (cGANs). When training cGANs, the generator takes in both a one-hot label vector, which describes the category that is to be generated, and a vector of random values drawn from some prior distribution, $p_z$. An improvement to training cGANS was proposed by Chen et al. \cite{chen2016infogan}; by seeking to maximise the mutual information between the one-hot label vector and the generated sample given the one-hot vector, the cGAN is encouraged to use the class label in the one-hot label vector, information which was often ignored in previous approaches \cite{mirza2014conditional, gauthier2014conditional}. We continue to use GANs in an unsupervised setting assuming no labels during training of the GANs. 

Adversarial training has also been applied to generative auto-encoders to impose a prior distribution on their encoding vectors \cite{makhzani2015adversarial}. For example, if the data distribution belongs to that of an ensemble of images, auto-encoders may be trained to compress each image to an encoding vector using an encoder and then to reconstruct an approximation of the same image from the encoding vector by using a decoder. Vectors may be passed into the decoder to generate new images, however the images will only be realistic if the input vectors to the decoder are from the same distribution as the encoding vectors of the ensemble. By imposing a prior distribution on the encoding during auto-encoder training, new ``encoding'' vectors may be drawn from the prior and passed through the decoder to generate new, meaningful image samples.  Makhzani et al. \cite{makhzani2015adversarial} showed that adversarial training is both better able to impose prior distributions and is able to impose more complex prior distributions.


GANs have not only be used for generation, but the representations learned during training have also been applied to discriminative tasks \cite{radford2015unsupervised,chen2016infogan,ajakan2014domain,creswell2016adversarial,donahue2016adversarial,dumoulin2016adversarially}. Since GANs are able to learn representations from unlabelled data, \cite{creswell2016adversarial, radford2015unsupervised}, they can be useful for learning representations when labelled data is not available, or the amount of labelled data is limited.

Until recently, representations used in discriminative tasks were obtained from trained GANs by passing samples through various layers of a trained discriminator, $D$. However, both Makhzani et al. \cite{makhzani2015adversarial} and Dumoulin et al. \cite{dumoulin2016adversarially} presented an alternative method for obtaining representations for samples by mapping image samples, $x$, back to $z$-space using an encoder which, under certain conditions, inverts the generator. This approach requires training an extra encoding network, and, in practice, this network often only approximately inverts the generator. In our work we continue to use the encoding from the discriminator so as to make minimal changes to the current adversarial network architecture. However, we would consider using encoding networks in future work.

A further application of adversarial training to representation learning is domain adaption, which involves learning a single representation for samples across different domains e.g. sketches and natural images. Both Ganin et al. and Ajakan et al. \cite{ganin2016domain, ajakan2014domain} apply adversarial training to learn representations for similar objects from different domains such that the representation for one domain cannot be distinguished from that of the other domain. The representations that are learned in this way may be applied to classification.  

\begin{figure}
    \centering
    \includegraphics[width=\columnwidth]{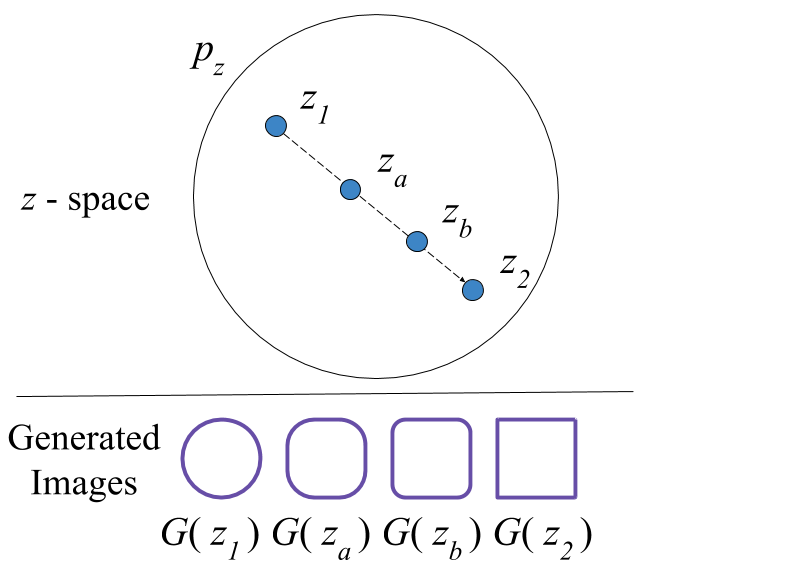}
    \caption{Interpolation is performed between two image samples $G(z_1)$ and $G(z_2)$ by generating images along a trajectory in $z$-space that lies between $z_1$ and $z_2$.}
    \label{interp}
\end{figure}

Despite the success of GANs as both generative and discriminative models, there are several problems that may still be addressed. 
For example, Radford et al. \cite{radford2015unsupervised} showed examples of interpolations between two random images, by generating images along a trajectory in $z$-space, see Fig. \ref{interp}. Often, samples towards the centre of the interpolation are poor, suggesting that the model $Q$ is assigning higher probability to regions where probability should be lower. An ideal generator, $G(z)$, should generate realistic samples for any $z \sim p_z$.

Further, previous work with GANs has ignored the implications of using a representation learned by a generative model for discriminative tasks \cite{radford2015unsupervised, dumoulin2016adversarially, makhzani2015adversarial,creswell2016adversarial, makhzani2015adversarial, salimans2016improved}. A representation learned using GANs tends to capture only a few regions of high density in $P$, failing to capture the whole data distribution \cite{huszar2015not,theis2015note,radford2015unsupervised}, which is not ideal for a representation that is intended for discriminative tasks: the representation is unlikely to generalise well to unseen samples. Failure of a GAN to capture the whole data distribution is evident when GANs generate similar samples for different $z$ inputs \cite{salimans2016improved}. Salimans et al. \cite{salimans2016improved} address the problem of $Q$ failing to capture more of the data distribution, $P$ by introducing ``mini-batch discrimination" which provides the discriminator with information about all samples in a batch to prevent similar samples being generated. Their approach is based only on heuristics. 
Salimans et al. \cite{salimans2016improved} found that employing ``mini-batch discrimination" led to the learning of a representation that performed better on discriminative tasks. This is consistent with our argument: a model that captures the whole data distribution should have an improved ability to generalise to new concepts, allowing representations extracted from such a model to be useful for discriminative tasks.


We aim to address both of these problems by providing a single, novel cost function, parametrised by $\pi:0 \le \pi \le 1$, that can be tuned to be more preferable for either generative or discriminative tasks.

Our alternative cost function can be tuned for generation by using a large $\pi$, which approximates the non-symmetric $KL[Q\|P]$. Minimising $KL[Q\|P]$ penalises the model, $Q$, when generated samples, $x \sim Q$, do not come from the training data distribution, $P$. By doing this, we increase the likelihood that samples drawn from our model are consistent with real samples; we find that our alternative cost function prevents nonsensical images being generated when interpolating between two random images using their $z$ space representations.

To tune our cost function for discriminative tasks, a small $\pi$ value may be used, which approximates the non-symmetric $KL[P\|Q]$. Minimising $KL[P\|Q]$ penalises the model, $Q$, when it does not capture all regions of density in $P$. We provide experimental evidence to show that this alternative cost function improves performance on several discriminative tasks including one-shot learning and retrieval when compared to regular GAN training.

\section{Proposed Cost Function}

The original cost function proposed by Goodfellow et al. \cite{goodfellow2014generative} is: \[\min_G \max_D V(G,D)= \mathbbm{E}_{x\sim P} \log D(x) + \mathbbm{E}_{z\sim p_z} \log (1 - D(G(z))) \]
We propose the alternative cost function:  \[\min_G \max_D V(G,D)= \pi  \mathbbm{E}_{x\sim P} \log D(x) +\]\[ (1-\pi) \mathbbm{E}_{z\sim p_z} \log (1 - D(G(z))) \]

We now show that under similar conditions and assumptions to those made by Goodfellow et al. \cite{goodfellow2014generative}, this new cost function is approximately proportional to $KL[P\|Q]$ and $KL[Q\|P]$ for large and small $\pi$ respectively.

\subsection{Proposed Cost Function In The Limits Of $\pi$}

First we show that for a fixed generator, $G_0(x)$, there exists an optimal discriminator, $D^*(x)$:
\[ D^*(x) = \max_D V(G_0,D) = \min_D - V(G_0,D) \]

\[V(G_0,D) = \pi  \mathbbm{E}_{x\sim P} \log D(x) +\]\[ (1-\pi) \mathbbm{E}_{z\sim p_z} \log (1 - D(G_0(z))) \]
\[= \pi \int_x P(x) \log D(x) dx + (1-\pi) \int_z p_z(z) \log (1-D((G_0(z))) dz \]
\[ = \pi \int_x P(x) \log D(x) dx + (1-\pi) \int_x Q(x) \log (1-D(x)) dx \]
\[ = \pi \int_x P(x) \log D(x) dx + (1-\pi) Q(x) \log (1-D(x)) dx \]

where $Q$ is the distribution of samples generated by $G(z)$.

To find the stationary curve, $D^*(x)$, of an integral over $x$, we use the Euler-Lagrange theorem.
For the general variational problem:
\[ \min_D I(u(x)), I(u(x)) = \int_a^b F(u(x),u'(x),x) dx\]\[ u(a)=u_a, u(b)=u_b \]
any differentiable and bounded minimiser, $u_0(x)$, is a solution to the boundary value problem:
\[S_F(x,u(x),u'(x)) \big|_{u(x)=u_0(x)} = \]\[ \frac{d}{dx} \left ( \frac{\partial F}{\partial u'(x)} \right )- \frac{\partial F}{\partial u(x)}\bigg|_{u(x)=u_0(x)}=0 \]
\[ \forall x \in (a,b), u(a)=u_a, u(b)=u_b \]

In the case where the integrand does not contain a $u'$ term, the boundary value problem simplifies to:
\[S_F(x,u(x)) \big|_{u(x)=u_0(x)} = \frac{\partial F}{\partial u}\bigg|_{u(x)=u_0(x)} = 0\]
implying that the stationary curve of the integrand is also the stationary curve of the integral.

Then, for any $(a,b) \in R^2 \notin \{0,0\}$,  the function $f(y)= a \log(y)+b \log(1-y)$ achieves a maximum in the interval [0,1] at $\frac{a}{a+b}$. So, we get:
\[D^*(x)=\frac{\pi P(x)}{\pi P(x) + (1-\pi)Q(x)}\]
Note that because the discriminator takes samples either from $P$ or $Q$, it is only defined for the $\supp(P(x)) \cup  \supp(Q(x))$, and so $P$ and $Q$ do not simultaneously equal zero, satisfying the conditions of $(a,b)$.

If the generator and discriminator are trained iteratively \cite{goodfellow2014generative}, one may assume that the discriminator is optimised in the first step of an iteration, giving the new cost function for the second step of the iteration, $C(G)=V(G,D^*)$: \[C(G)=\pi \mathbbm{E}_{x\sim P} \log\left (\frac { \pi P(x)}{\pi P(x)+ (1-\pi) Q(x)}\right )\]\[ + (1-\pi) \mathbbm{E}_{x\sim Q} \log \left (\frac {(1- \pi) Q(x)}{\pi P(x)+ (1-\pi) Q(x)}\right )\]
Which can be re-arranged to give:
\[C(G)= \pi \log \pi + (1-\pi) \log (1-\pi)\]\[ + \pi KL[P \| \pi P + (1-\pi) Q] + (1-\pi) KL[Q|| \pi P + (1-\pi) Q]\]
\[ = \pi \log \pi + (1-\pi) \log (1-\pi) + JS_{\pi} [P \| Q] \]

Now, we consider the limits as $\pi \rightarrow 0$ and $\pi \rightarrow 1$, knowing that $JS_\pi$ is proportional to $KL[P\|Q]$ for small $\pi$ and proportional to $KL[Q\|P]$ for large $\pi$ \cite{huszar2015not}:
\[ \lim_{\pi \rightarrow 0}  \pi \log \pi + (1-\pi) \log (1-\pi) + JS_{\pi} [P \| Q]  \]
\[= 0 + \lim_{\pi \rightarrow 0} JS_{\pi} [P \| Q] \propto KL[P\|Q] \]
and
\[ \lim_{\pi \rightarrow 1}  \pi \log \pi + (1-\pi) \log (1-\pi) + JS_{\pi} [P \| Q]  \]
\[= 0 + \lim_{\pi \rightarrow 1} JS_{\pi} [P \| Q] \propto KL[Q\|P] \]

We have shown that the cost function that we propose approximates $KL[P\|Q]$ for small $\pi$ and approximates $KL[Q\|P]$ for large $\pi$. Which implies that to train a model, $Q$, suitable for retrieval, our proposed cost function can be used with a small $\pi$, and to train a model suitable for generation our proposed cost function can be used with a larger $\pi$ value. We explore the practical implications of this in the context of retrieval, generation and one-shot learning in the experimental section.

\section{Experiments \& Results}
There are currently two main types of application for generative models. The first is the synthesis of novel samples that resemble the training data, and the second is to for discriminative tasks, such as classification and retrieval. The latter make use of the representation that is learned by a network during generative training.

In this section, we evaluate our alternative cost function on three tasks: 
\begin{itemize}
    \item Generation of novel samples
    \item One-shot classification
    \item Retrieval of visually similar samples
\end{itemize}

We compare the performance of GANs trained using the alternative cost functions of these tasks for $\pi \in \{0.01, 0.5, 0.9\}$; a consideration of the limiting values of $\pi$ suggests that a model trained using small values ($e.g. \pi=0.01$) should perform better on retrieval and one-shot classification tasks, whilst a model trained using, say, $\pi=0.9$ should perform better on generative tasks. The purpose of these experiments is to provide experimental evidence to support this, based on the analysis of Section IV.

\begin{figure}
    \centering
    \includegraphics[width=\columnwidth]{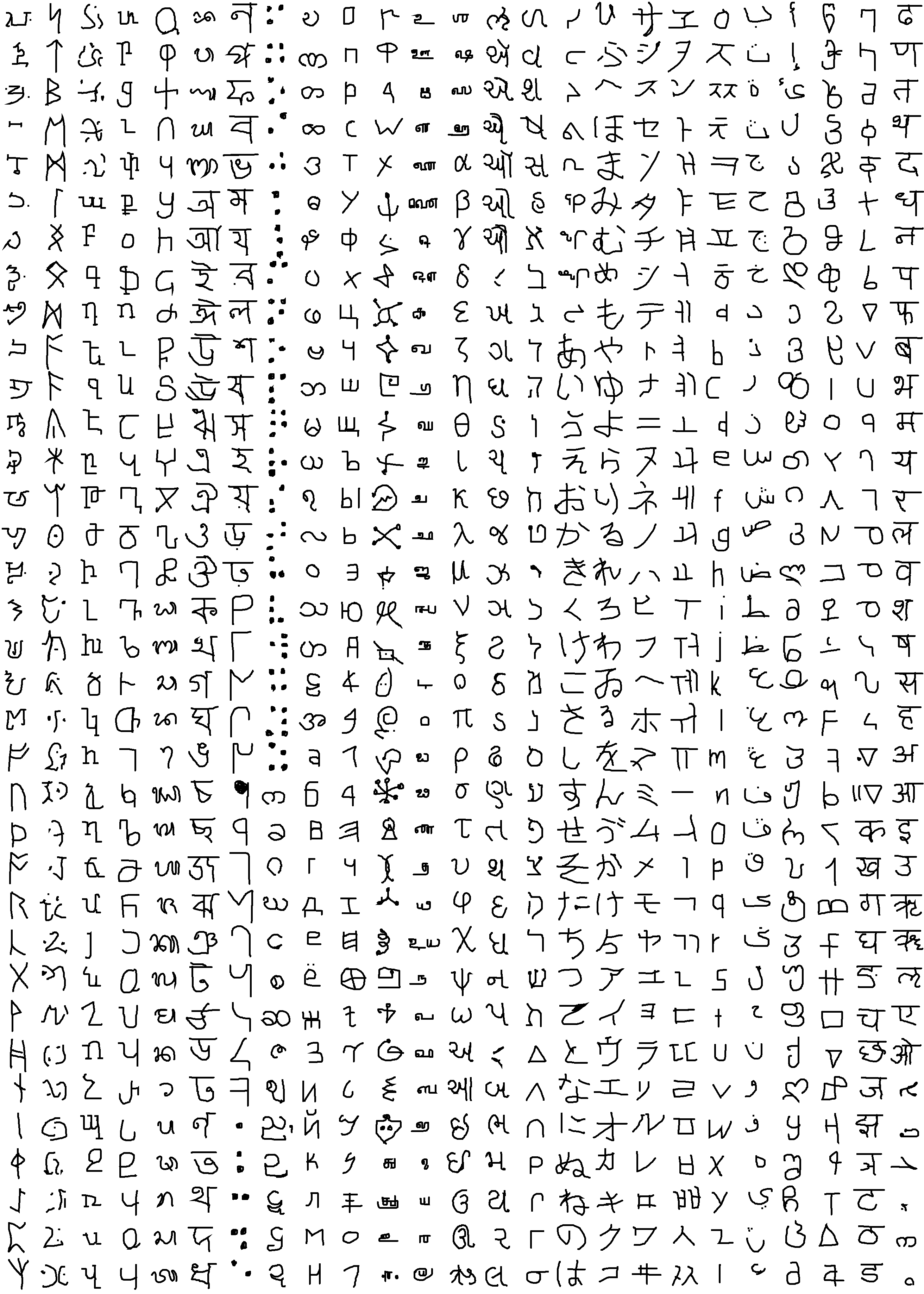}
    \caption{Examples of hand-written characters from the Omniglot background dataset \cite{lake2015human}.}
    \label{omni_back_montage}
\end{figure}

\subsection{Dataset}
We apply our alternative cost function to the Omniglot dataset \cite{lake2015human}, see Fig. \ref{omni_back_montage}.
Previously, generative adversarial networks have been trained on handwritten numbers (MNIST), street numbers (SVHN), faces (CelebA) and natural scenes (CIFAR10).

\begin{figure}
    \centering
    \includegraphics[width=\columnwidth]{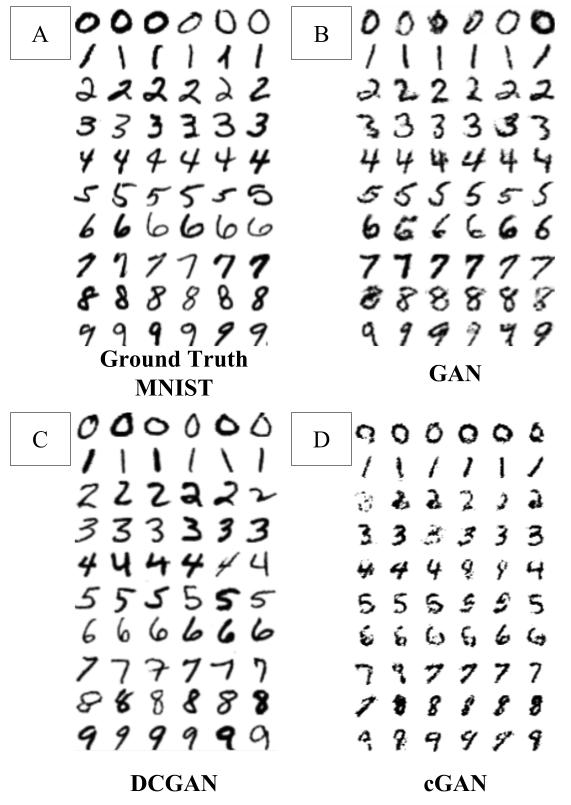}
    \caption{Previous work using GANs to generate hand-written digits. A) Shows examples of the MNIST samples used to train the GANs in B-D, B) Generations using fully connected GANs \cite{goodfellow2014generative}, C: Improved generations using deep convolutional generative adversarial networks \cite{radford2015unsupervised}. (A-C: Images modified from \cite{goodfellow2014generative}). and D: Conditional generations using conditional GANs : Image modified from \cite{mirza2014conditional}. }
    \label{MNIST_GAN_eg}
\end{figure}

Once trained, the generator of a GAN \cite{radford2015unsupervised} is able to generate hand-written digits that are indistinguishable from real samples, see Fig. \ref{MNIST_GAN_eg}C. Hand-written digits generated by the trained generator of a conditional GAN \cite{mirza2014conditional} are recognisable as numbers, see Fig. \ref{MNIST_GAN_eg}D. On the other hand, generation of realistic looking natural image scenes has not yet been achieved. The MNIST dataset consists of only 10 classes each with $60,000$ examples in total. In contrast, the Omniglot \cite{lake2015human} dataset has $1,623$ classes with only $20$ examples of each class. For all of our experiments, we use the Omniglot dataset for several reasons:
\begin{itemize}
    \item The Omniglot dataset is neither as simple as the MNIST dataset, nor as complex as the CIFAR-10 dataset, which means improvements to regular GAN training may be more evident.
    \item There are only a few labelled examples per class, which makes the Omniglot dataset a perfect candidate for using adversarial training to learn a representation for discriminative tasks.
\end{itemize}

The Omniglot dataset \cite{lake2015human} contains characters from $50$ different writing systems. The dataset is split into a background dataset of $964$ characters from $30$ writing systems, while the evaluation dataset consists of $659$ different characters from $20$ different writing systems. A GAN is trained on the background dataset, using our proposed alternative cost function with each $\pi$ value. Note that although the dataset has labels, the labels are not used at any point during training.

\begin{algorithm}
\begin{algorithmic}
    \FOR {Number of training iterations}
    \FOR {$k$ iterations}
    \STATE $z_1 ... z_m \sim p_z$ \textit{ \# Get m samples from the prior}
    \STATE $x_1 ... x_m \sim P$  \textit{ \# Get m samples from the data}
    \STATE \textit{\# Calculate discriminator loss:}
    \STATE \textit{\# \textbf{using our proposed alternative cost function}}
    \STATE $J_D = -\frac{1}{2m}\big( \pi \sum_{i=1}^m \log D(x_i) + $
    \STATE $(1-\pi)\sum_{i=1}^m \log (1-D(G(z_i))) \big )$
    \STATE $\theta_G \leftarrow \theta_G - \nabla_{\theta_G} J_G$ \textit{ \# Update weights}
    \ENDFOR 
    \STATE $z_1 ... z_m \sim p_z$ \textit{ \# Get m samples from the prior}
    \STATE \textit{ \# Calculate the generator error}
    \STATE $J_G=-\frac{1}{m} \big(\sum_{i=1}^m \log (D(G(z_i))) \big)$
    \STATE $\theta_G \leftarrow \theta_G - \nabla_{\theta_G} J_G$ \textit{ \# Update weights}
    \ENDFOR
\end{algorithmic}
\caption{Algorithm For Training a GAN: Similar to the training algorithm of Goodfellow et al. \cite{goodfellow2014generative} but incorporating the proposed change to the cost function.}
\label{train_alg}
\end{algorithm}

\begin{figure}
    \centering
    \includegraphics[width=\columnwidth]{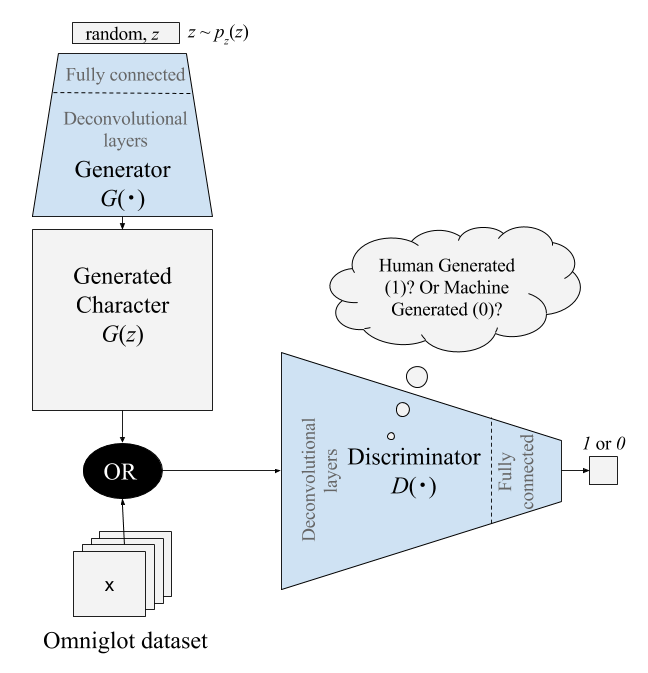}
    \caption{GAN architecture: Fig \ref{gen_model} gives a conceptual model for how a GAN works. Here we present the overall GAN architecture. A random sample, $z$ is drawn from a prior distribution $p_z(z)$ and passed through the generator to generate an image. The generator consists of a fully connected layer and a series of deconvolutional layers. An image either from the generator or the training dataset is passed through the discriminator to predict if the image was from the training data or not. The discriminator consists of a series of convolutional layers and a fully connected layer.  Details of both the generator and discriminator architecture can be found in Table \ref{arch}.}
    \label{omni_GAN}
\end{figure}

\subsection{Architecture \& Training}
For training purposes, the generator, $G$, and discriminator, $D$, may be any differentiable functions; here we used deep convolutional neural networks, see Fig. \ref{omni_GAN}.
The $D$ network is a regular feed forward convolutional neural network. As suggested by Radford at al. \cite{radford2015unsupervised}, we used convolutions applied with stride two \cite{ springenberg2014striving,long2015fully} to down-sample the image instead of using pooling.
The $G$ network requires upsampling, which cannot be achieved by a regular feed forward network. One method that can be used to upsample the images appropriately would be to use the error tensor (gradient image) for a convolution layer applied with a stride of two. However, we simply applied filters via convolution with stride one and upsampled the resulting image array using bilinear interpolation.
The architecture of the networks is similar to that of Radford et al. \cite{radford2015unsupervised}. The training images used by Radford et al. \cite{radford2015unsupervised} were $64 \times 64$ compared to the Omniglot images which are $105 \times 105$. To account for this difference in shape, the fully connected layers of both $G$ and $D$ have more nodes, so that the size of the activation images entering the first convolutional layer in $G$  are of size $13 \times 13$ instead of $4 \times 4$. Another modification is the size of the filters in the final layer of $G$: we used filters of size $4 \times 4$ to accommodate for the output image having odd-valued dimensions.
All networks were initially trained for $2,000$ iterations using random batches of $128$ samples with learning rate of $0.002$, a $10 \times$ faster learning rate than that of Radford et al. \cite{radford2015unsupervised} and a $k$ value of 3. However, we found that for $\pi=0.9$, the network did not converge after $2,000$ iterations. Instead, we trained with a $k$ value of 1 for $\pi=0.9$. The latent variable, $z$, that is the input to the generator, has dimension $n=100$, and is drawn from a uniform distribution, $U[0,1]$.

\setlength{\tabcolsep}{4pt}
\begin{table}
\begin{center}
\caption{Network Architecture Used. FC=fully connected layer, C=convolutional layer with stride 2, D=convolutional layer with stride 0.5, unless stated otherwise. For all experiments in this paper, $n$=100.  ``batch norm.'' refers to batch normalisation.}
\label{arch}
\begin{tabular}{llll}
\hline\noalign{\smallskip}
$G$ & $D$\\
\noalign{\smallskip}
\hline
\noalign{\smallskip}
FC: $43264 \times n $, reshape(256,13,13)& C: $32 \times 1 \times 5 \times 5$ \\ 
batch norm., leakyReLU(0.2)     & batch norm., ReLU \\
D: $128 \times 256 \times 5 \times 5$   & C: $128 \times 32 \times 5 \times 5$ \\
batch norm., leakyReLU(0.2)     & batch norm., ReLU \\  
D: $64 \times 128 \times 5 \times 5$    & C: $256 \times 128 \times 5 \times 5$ \\
batch norm., leakyReLU(0.2)     & batch norm., ReLU,\\
& reshape(50176)\\
D: $1 \times 64 \times 4 \times 4$      & FC: $50176 \times 1$\\

\hline
\end{tabular}
\end{center}
\end{table}
\setlength{\tabcolsep}{1.4pt}

\subsection{Retrieval}
The Omniglot dataset consists of a background and an evaluation set. The background set consists of characters from different alphabets to the evaluation dataset. A GAN is trained on the background dataset using $\pi \in \{0.01, 0.5, 0.9\}$, where training with $\pi=0.5$ is equivalent to normal GAN training. Retrieval is performed both on the background and the evaluation dataset. To retrieve examples of characters not seen before, the representation that is learned during training should capture the entire distribution of handwritten space in order to generalise well to new concepts. We expect a GAN that is trained using $\pi=0.01$ to outperform regular GAN training ($\pi=0.5$), since a GAN trained using $\pi=0.01$ approximately minimises $KL[P\|Q]$ thus encouraging the model $Q$ to capture more of the data distribution. By contrast, we expect a GAN trained using $\pi=0.9$ to perform worse than regular GAN training on the evaluation dataset: a GAN trained using $\pi=0.9$ approximately minimises $KL[Q\|P]$, encouraging the model $Q$ to only capture the densest parts of the training data distribution. Such a model would not be expected to generalise well to unseen parts of the distribution. However, when retrieving from the background dataset, it is likely that for $\pi=0.9$, retrieval will be similar to that of $\pi=0.5$ because the model does not have to generalise to new concepts well, as training and testing are performed on the same dataset.


To perform retrieval, both a query sample and samples in the retrieval dataset (either background or evaluation) are encoded. To encode a sample, it is passed through to the penultimate layer of the discriminator to give a $50$k dimensional encoding vector. The cosine similarity measure is calculated between the query and all samples in the retrieval dataset to score their similarity. The most similar matches are returned in descending order of similarity.

\subsubsection{Retrieval Across Multiple Alphabets}
\begin{figure}
    \centering
    \includegraphics[width=0.95\columnwidth]{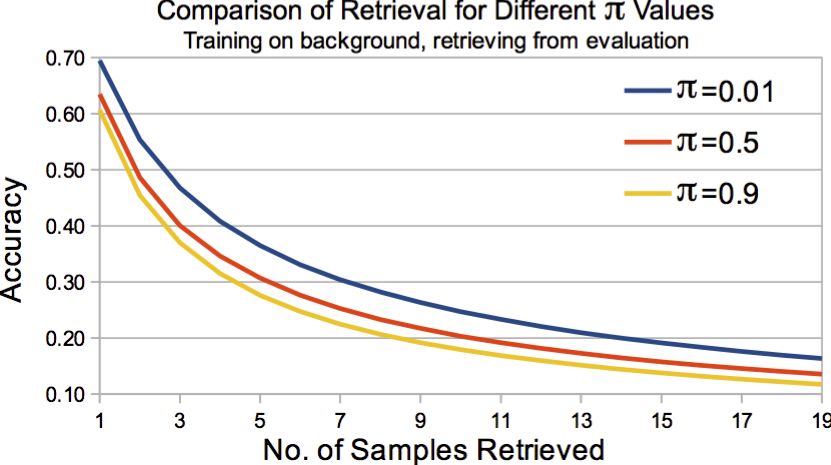}
    \caption{Comparing accuracy vs. retrieval on the Omniglot evaluation dataset for regular GAN training ($\pi=0.5$) and our alternative cost function using $\pi \in \{ 0.01, 0.9\}$.}
    \label{PR_omni_eval}
\end{figure}

For each query character in the Omniglot dataset, there are $19$ similar examples, so  we retrieve the top $19$ matches for any query from the evaluation dataset. 
We treat every sample in the evaluation dataset, in turn, as a query and take the average accuracy across queries. Fig. \ref{PR_omni_eval} shows the average accuracy-retrieval curve for the top $19$ retrievals across all queries. As expected, using $\pi=0.01$ improved retrieval compared to regular GAN training ($\pi=0.5$) while $\pi=0.9$ worsened retrieval compared to regular GAN training. 

\begin{figure}
    \centering
    \includegraphics[width=\columnwidth]{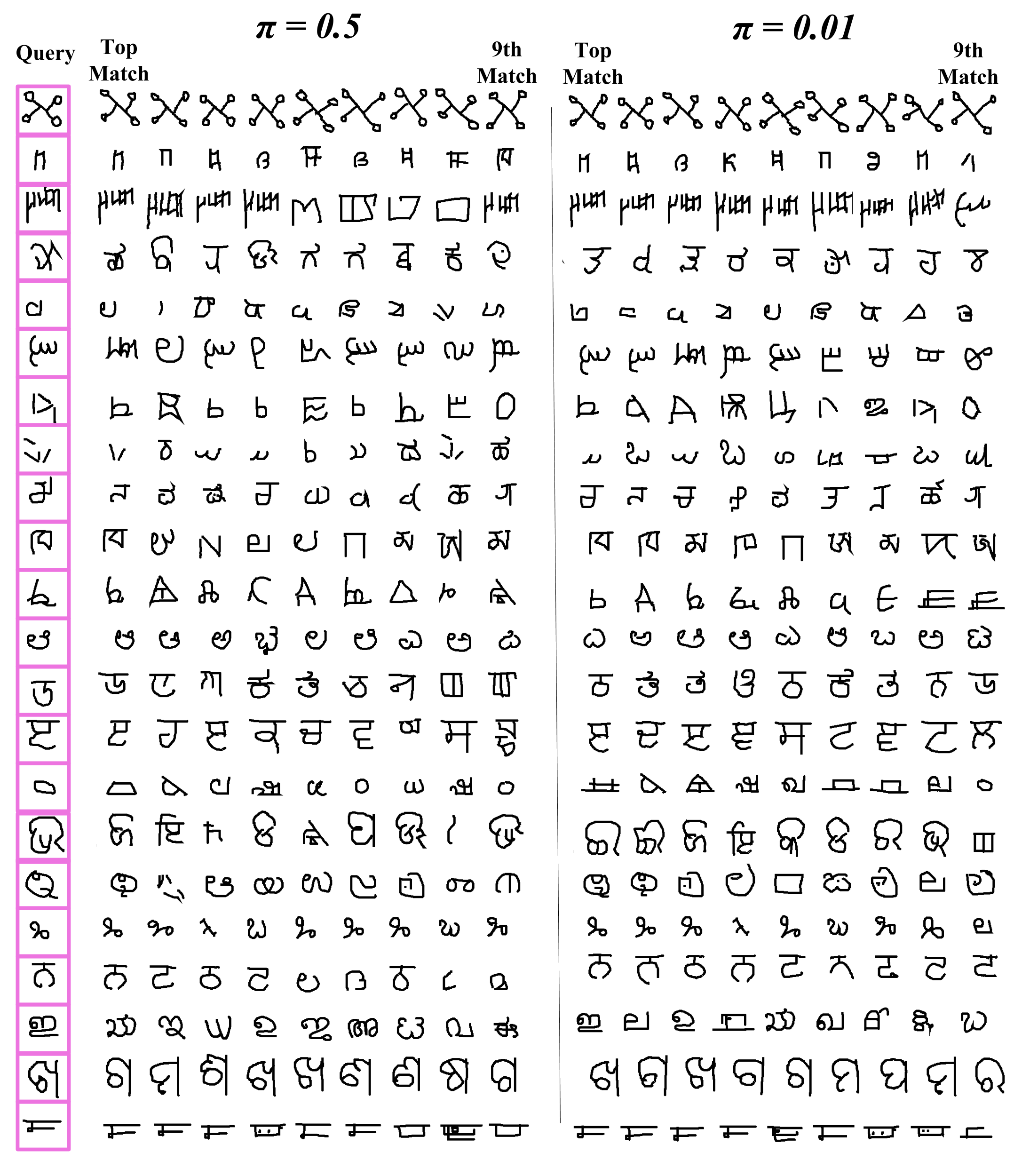}
    \caption{Comparing top $9$ retrievals on the Omniglot evaluation dataset for regular GAN training ($\pi=0.5$) and our alternative cost function using $\pi=0.01$. }
    \label{omni_ret}
\end{figure}

For the task of retrieval, we are particularly interested in how setting $\pi=0.01$ improves performance. Fig. \ref{omni_ret} shows the top $10$ retrievals using $\pi=0.5$ and $\pi=0.01$ for a selection of queries. Using $\pi=0.01$, the accuracy of the top retrieval is $69.48\%$ achieved on the evaluation dataset compared to $63.51\%$ when using $\pi=0.5$. The chance of randomly retrieving a matching sample is $0.15\%$, a $\pi$ value of $0.01$ improves top-1 retrieval accuracy by nearly $40$ times that of chance.

\begin{figure}
    \centering
    \includegraphics[width=\columnwidth]{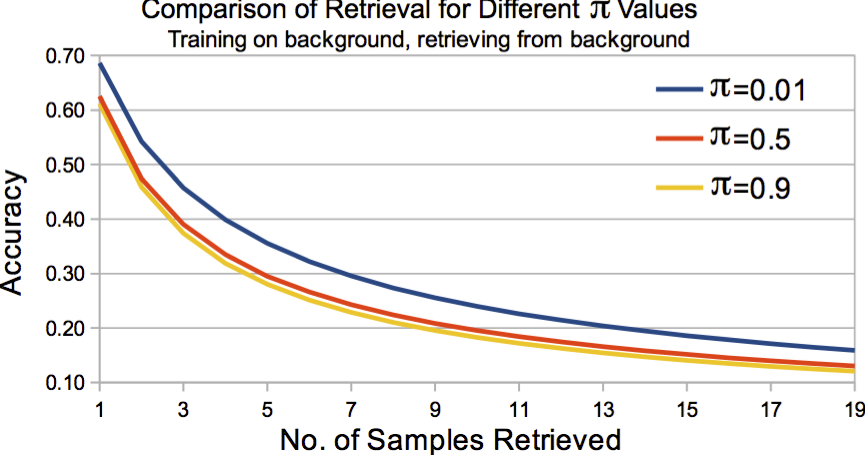}
    \caption{Comparing accuracy vs. retrieval on the Omniglot background dataset for regular GAN training ($\pi=0.5$) and our alternative cost function using $\pi \in \{0.01, 0.9\}$.}
    \label{PR_omni_back}
\end{figure}

When retrieving from the background dataset using $\pi=0.5$, the accuracy of the top retrieval is $62.51\%$ compared to $68.59\%$ when using $\pi=0.01$. The chance of randomly retrieving a matching sample is $0.1\%$, so here a $\pi$ value of $0.01$ improves accuracy performance by a factor of $60$ relative to random choice. The accuracy-retrieval curve can be see in Fig. \ref{PR_omni_back}. A summary of results is shown in Table \ref{comparison}.

\setlength{\tabcolsep}{4pt}
\begin{table}
\begin{center}
\caption{Comparison of retrieval accuracy on the evaluation set training using our alternative cost function with different $\pi$ values. Note that $\pi=0.5$ is equivalent to regular GAN training.}
\label{comparison}
\begin{tabular}{llll}
\hline\noalign{\smallskip}
Method & Accuracy (Top 1) \\
\noalign{\smallskip}
\hline
\noalign{\smallskip}
GAN $\pi=0.9$ & $60.70\%$ \\
GAN $\pi=0.5$ & $63.51\%$ \\
GAN $\pi=0.01$ & $69.48\%$\\
\hline
\end{tabular}
\end{center}
\end{table}
\setlength{\tabcolsep}{1.4pt}



\subsubsection{Retrieval Within Alphabets}
\begin{figure}
    \centering
    \includegraphics[width=\columnwidth]{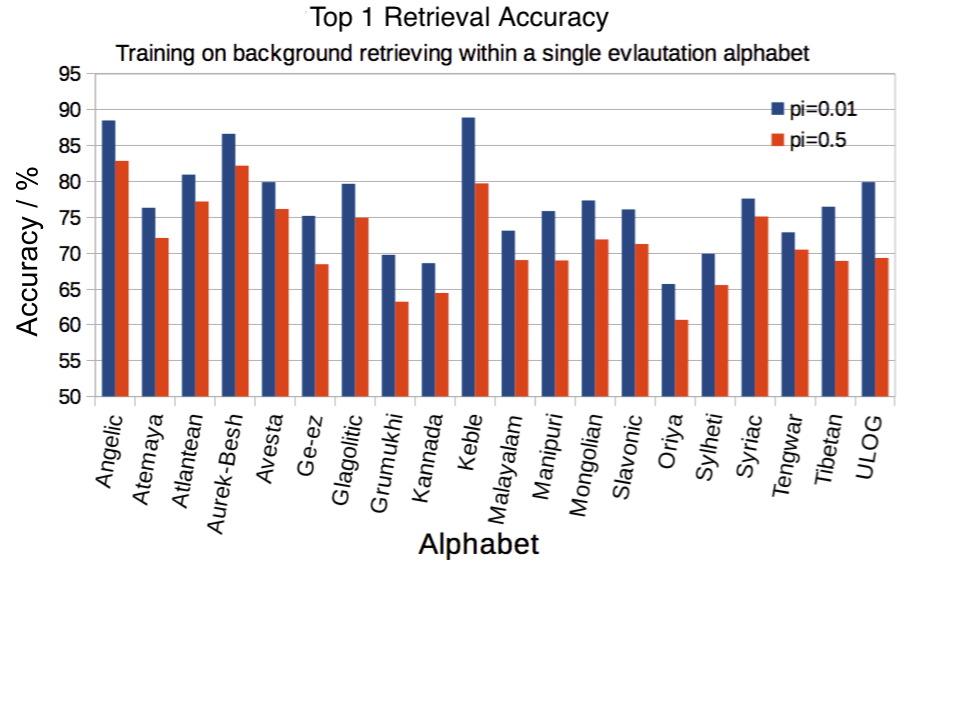}
    \caption{Comparing top 1 retrieval accuracy for each alphabet in the evaluation set for regular GAN training ($\pi=0.5$) and our proposed alternative cost function using $\pi=0.01$.}
    \label{ret_ABC}
\end{figure}

We also apply our proposed system to perform retrieval on individual alphabets, and compare GANs trained using our alternative cost function at $\pi=\{0.5, 0.01\}$ where again, according to the theory in Section IV, we expect training with $\pi=0.01$ to perform best. Fig. \ref{ret_ABC} shows the accuracy of the top retrieval on each alphabet for $\pi=0.5$ and $\pi=0.01$. Results show that a GAN trained using the alternative cost function with $\pi=0.01$ improves retrieval performance on all alphabets.

\subsection{One-Shot Classification}
Humans are often able to learn very quickly from only a few examples; training machines to learn from few examples is more difficult. The machine equivalent to learning from few examples is $K$-shot learning, where in the extreme case, $K=1$ and a classifier learns from only one example. Typically, classification models need to learn from many examples to capture the variation of samples in a dataset. Convolutional neural networks typically learn from millions of images \cite{krizhevsky2012imagenet}, making this task very challenging.

In these experiments, a representation for handwritten characters is learned by training two GANs with our alternative cost function. The first GAN is trained using $\pi=0.5$, equivalent to regular GAN training. The second is trained using $\pi=0.01$, which we would expect to learn a representation more suitable for one-shot learning.



Previous work \cite{vinyals2016matching, santoro2016one} has looked at learning labels for five or $20$ randomly chosen classes from the Omniglot dataset across all alphabets having been shown one or five examples during training. 


Learning to classify only five random samples from different classes across the dataset is of less practical significance compared to learning to classify all samples in the dataset or all samples within a single alphabet. For this reason, we perform the novel task of one-shot learning on both the whole dataset and on individual alphabets. These tasks are more challenging than those of Vinyals et. al \cite{vinyals2016matching} and Santoro et al. \cite{santoro2016one} for several reasons:
\begin{enumerate}
    \item By picking samples randomly across all alphabets, the chance of picking two samples from the same alphabet is minimised. Samples from within an alphabet often bear greater similarity to each other than samples from different alphabets, making it easier to perform one-shot learning across alphabets than within alphabets.
    \item Each alphabet in the evaluation dataset has between $20$ and $55$ character classes, which makes the classification task harder since the probability of randomly guessing the correct label is smaller.
    \item For training and testing, the dataset provided by Lake et al. \cite{lake2015human} is used; it is split into $964$ training classes and $659$ testing classes, while Vinyals et al. \cite{vinyals2016matching} and Santoro et al. \cite{santoro2016one} split the data into $1200$ training and $423$ testing.
\end{enumerate}

Vinyals et al. \cite{vinyals2016matching} and Santoro et al. \cite{santoro2016one} further boost performance by employing data augmentation methods which have been shown to improve classification results by preventing over fitting -- a common problem when the quantity of training data is limited. We do not use data augmentation since we wish to focus our evaluation of the quality of the representation that is learned by using different $\pi$ values.

\begin{figure}
    \centering
    \includegraphics[width=\columnwidth]{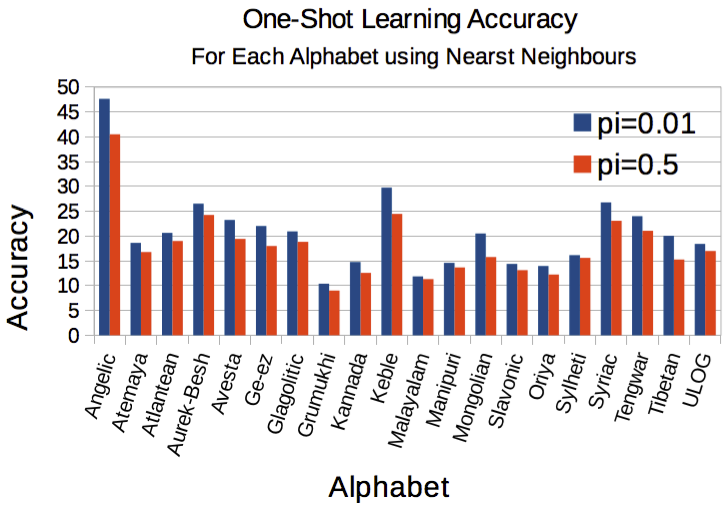}
    \caption{Comparing One-shot learning accuracy for regular GAN training ($\pi=0.5$) and our proposed alternative cost function using $\pi=0.01$.}
    \label{one_shot}
\end{figure}

Fig. \ref{one_shot} shows the results of one-shot classification on individual alphabets. A Nearest Neighbours classifier for each alphabet at each $\pi$ value was trained on a single sample from each character class within that alphabet, encoded using the discriminator of the GAN trained on the background dataset. The classifier was evaluated on the rest of the samples in the dataset to give the scores shown in Fig. \ref{one_shot}.

Results of one-shot learning for the entire evaluation dataset are shown in Table \ref{one_shot_eval}, training Nearest Neighbours (NN) and LinearSVM on a single example of each character {\em class} by encoding them using the trained discriminators at each $\pi$ value.

\setlength{\tabcolsep}{4pt}
\begin{table}
\begin{center}
\caption{Comparison of One-shot learning accuracy on the whole evaluation set training using our alternative cost function with different $\pi$ values, where $\pi=0.5$ is equivalent to regular GAN training.}
\label{one_shot_eval}
\begin{tabular}{p{2cm}p{2.5cm}p{2.5cm}}
\hline\noalign{\smallskip}
Method & Accuracy using $\pi=0.5$ & Accuracy using $\pi=0.01$ \\
\noalign{\smallskip}
\hline
\noalign{\smallskip}
$1$-NN & $7.25\%$ & $9.58\%$  \\
LinearSVM & $6.69\%$ & $9.31\%$ \\
\hline
\end{tabular}
\end{center}
\end{table}
\setlength{\tabcolsep}{1.4pt}


One-shot classifiers trained with features that have been taken from a trained GAN with $\pi=0.01$ outperforms classifiers trained with regular ($\pi=0.5$) GAN features on all the alphabets (Fig. \ref{one_shot}) and across the dataset as a whole (Table \ref{one_shot_eval}). This supports the assertion that alternative training with smaller $\pi$ values is better suited to discriminative tasks than regular training of GANs.

\subsection{Generating Image Samples}
Characters in the Omniglot dataset are made up of strokes \cite{lake2015human}, with  some characters having similar strokes to each other. The background dataset used for training the GAN consists of $946$ characters with $20$ examples per character, this means that the GAN has nearly $20$k examples to learn strokes from, but only $20$ examples to learn specific characters. 

\begin{figure}
    \centering
    \includegraphics[width=\columnwidth]{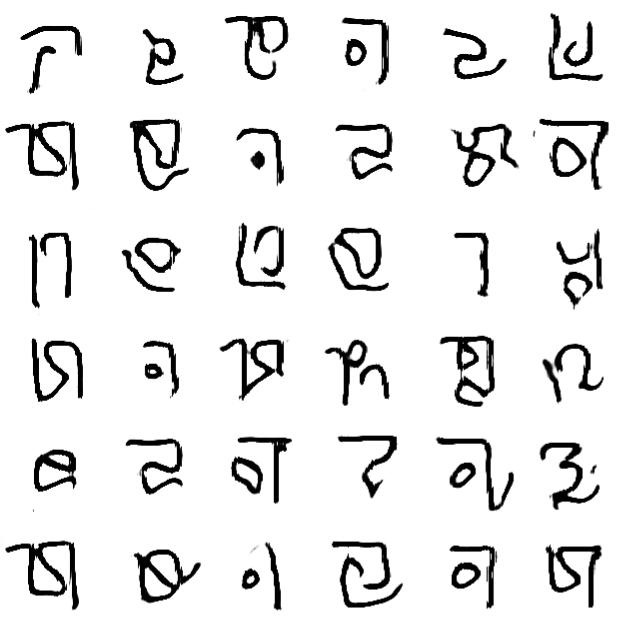}
    \caption{Omniglot random generations from a GAN trained using regular methods ($\pi=0.5$).}
    \label{omni_gen_0.5}
\end{figure}

\begin{figure}
    \centering
    \includegraphics[width=\columnwidth]{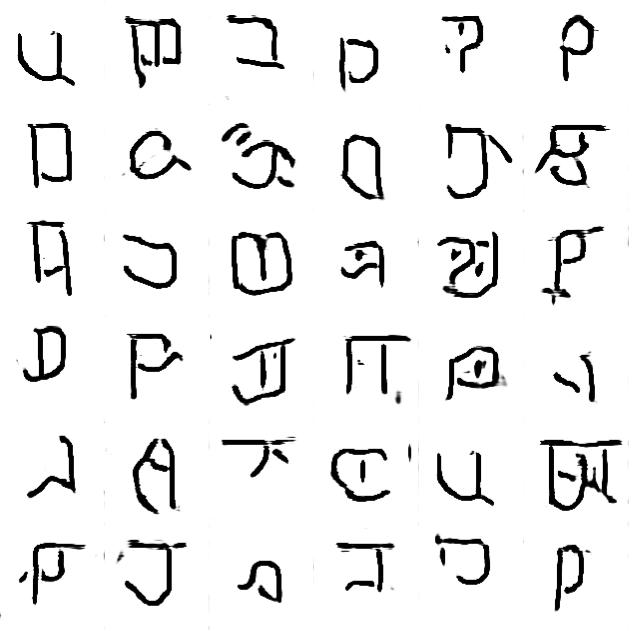}
    \caption{Omniglot random generations from a GAN trained using our alternative adversarial cost function with $\pi=0.9$.}
    \label{omni_gen_0.9}
\end{figure}

A GAN is trained using our alternative cost function with $\pi \in \{0.5, 0.9\}$. We generated $36$ random samples by drawing $36$, $100$-dimensional $z$ values from a uniform distribution and passing them through the trained generator. The results are shown in Fig. \ref{omni_gen_0.5} and Fig. \ref{omni_gen_0.9}. In comparing these two figures, it is difficult to draw any conclusions about any benefit to using a larger $\pi$ value; however, experiments involving interpolation show a clear distinction in the way that the generator captures the image space through $z$. This is explored in Section \ref{subsubsec:Interp}.

\subsubsection{Checking for over fitting}
\begin{figure}
    \centering
    \includegraphics[width=\columnwidth]{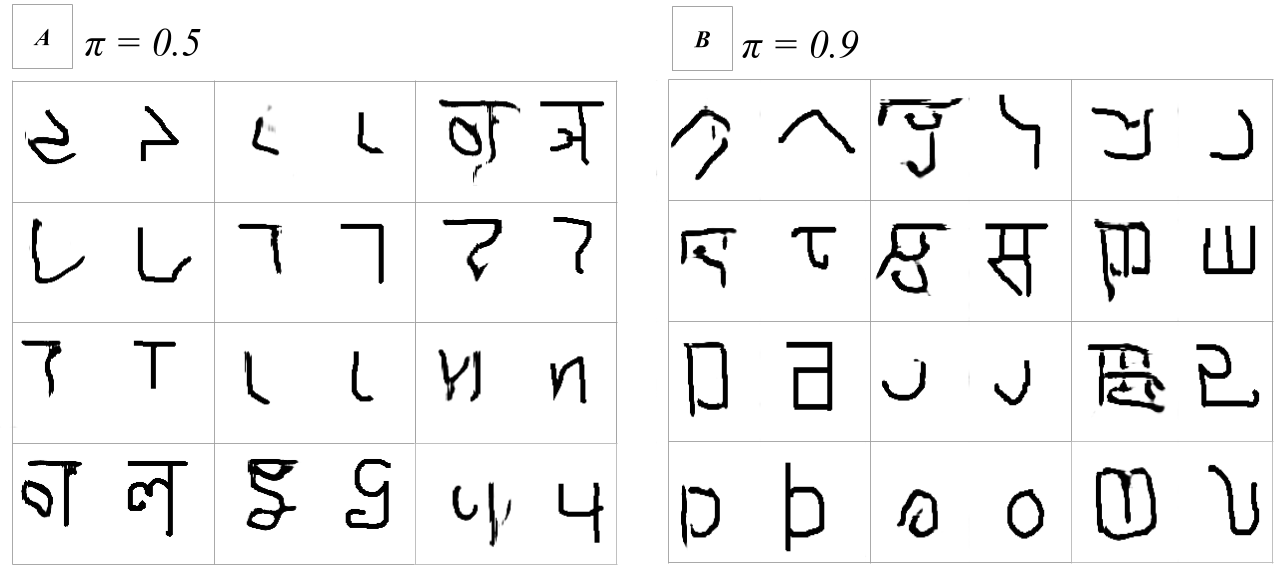}
    \caption{Pixel-wise nearest neighbour real samples to generated samples. A: For regular GAN training ($\pi=0.5$), B: Using our alternative cost function with $\pi=0.9$.}
    \label{omni_nearest}
\end{figure}

To show that our generator does not simply over fit to samples from the training data, we show, in Fig. \ref{omni_nearest}, examples of generated samples alongside their pixel-wise nearest neighbour sample from the training data. Results show that the generations are not exact copies of samples from the training data. Further, they strongly suggest that some of the image samples that are generated belong to character classes from the training dataset. However, to match generated samples to a character class, pixel-wise nearest neighbours might not be sufficient.

\subsubsection{Interpolating between random image samples}
\label{subsubsec:Interp}
The generator should generate realistic looking samples for any sample, $z$, drawn from the prior distribution, in this case a uniform distribution. According to the analysis in Section IV, training the networks with $\pi=0.9$ should encourage the generator to learn a model that captures only the densest parts of the training data distribution at the cost of ignoring the less dense regions. This suggests that samples drawn from a model trained using $\pi=0.9$ are more likely to give visually realistic samples than a model trained using $\pi=0.5$.

\begin{figure}
    \centering
    \includegraphics[width=\columnwidth]{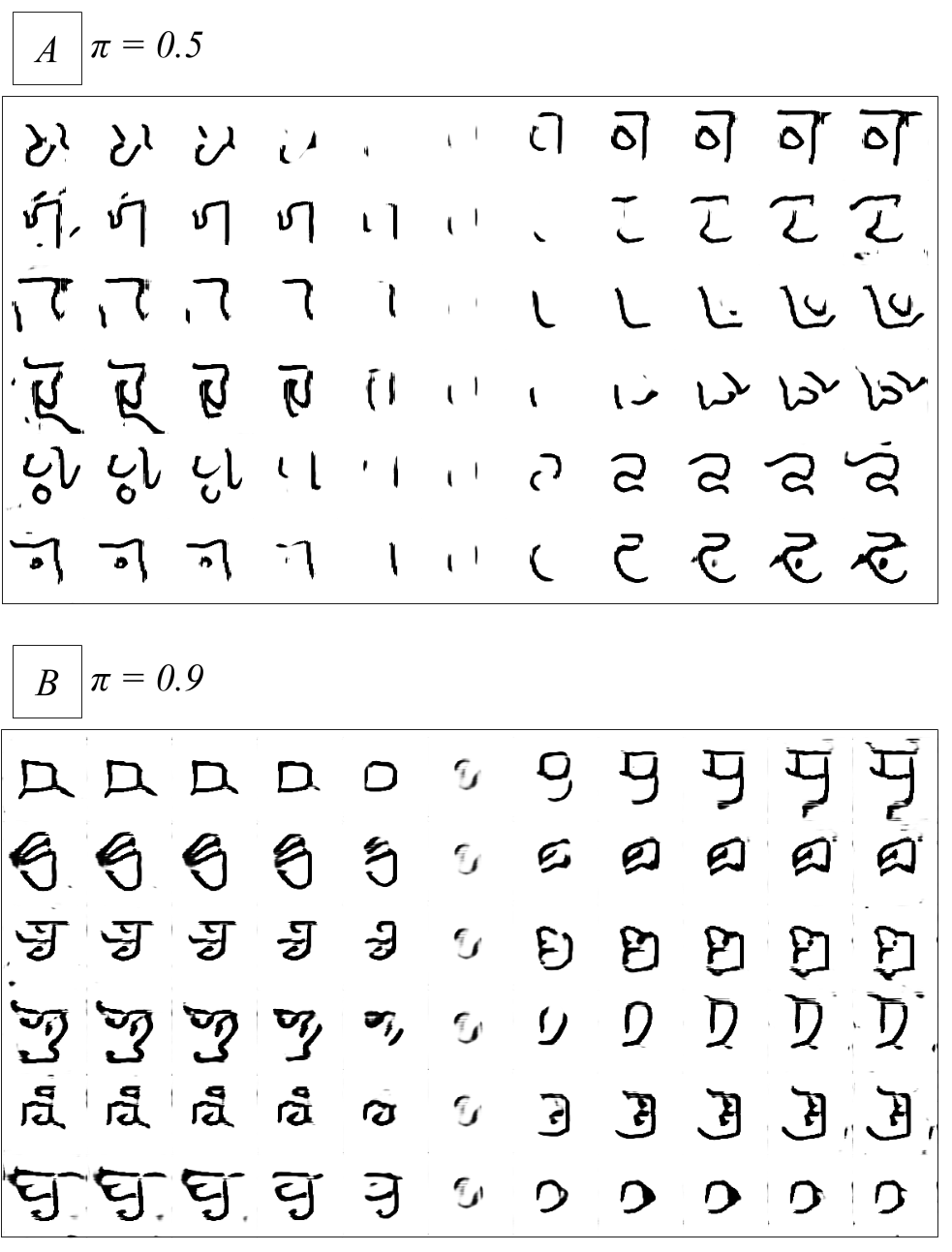}
    \caption{Comparing uniform interpolations in z-space between random start and end samples, A: For regularly trained GANs ($\pi=0.5$) and B: GANs trained using the alternative cost function with $\pi=0.9$}
    \label{omni_interp}
\end{figure}

To test this hypothesis, we would have to generate samples for all possible $z$, which is not feasible. Instead, we take two random $z$ values from the prior distribution and linearly interpolate between them at $9$ points and generate samples from these points. These are shown in Fig. \ref{omni_interp}. For both $\pi=0.5$ and $\pi=0.9$, the samples at the intermediate points appear to fail, particularly towards the centre of the interpolation. However, for $\pi=0.9$ the change is more abrupt and the first and last $4$ samples in the interpolations are consistently good, whereas only the fist and last $3$ are consistently good for $\pi=0.5$. 

\begin{figure}
    \centering
    \includegraphics[width=\columnwidth]{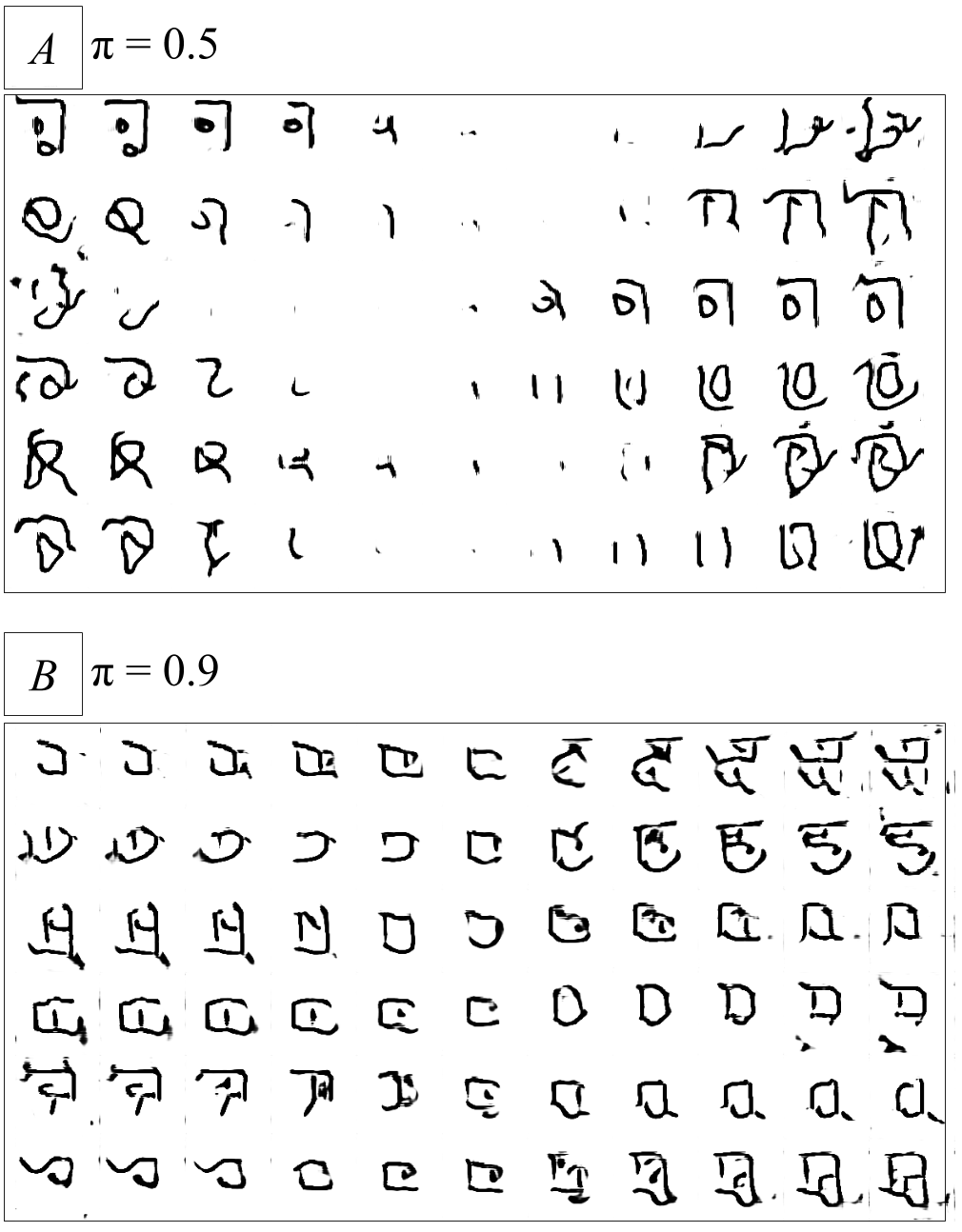}
    \caption{Comparing spherical interpolation in z-space between random start and end samples, A: for regulalrly trained GANs ($\pi=0.5$) and B: GANs trained using the alternative cost function with $\pi=0.9$. Note the apparent missing samples in A.}
    \label{omni_spherical_interp}
\end{figure}

Using linear interpolation in high dimensions often leads to taking uneven steps between the samples. An alternative interpolation that takes even steps between samples is spherical interpolation \cite{shoemake1985animating}, giving a more representative view of the space between samples. Fig. \ref{omni_spherical_interp} shows hyperspherical interpolations between random samples in $z$-space for $\pi=0.5$ and $\pi=0.9$. Here, the effect of $\pi$ is more evident. When evenly traversing $z$-space between two random samples, there are more nonsensical gaps when  samples are drawn from a GAN trained using $\pi=0.5$ than those drawn from a GAN trained using $\pi=0.9$. The results for $\pi=0.9$ are consistent with a model that has optimised $KL[Q\|P]$, to ensure that any sample drawn from the model is likely to come from the same distribution as the training data. This further supports the hypothesis that training a GAN using larger $\pi$ values is more suitable for sample generation.


\section{Discussion}
When showing that adversarial training is equivalent to $KL$ divergence for large and small $\pi$, it is assumed that $D$ is near-optimal. To improve the chance that $D$ is close to optimal, for every one iteration that we train $G$ for values of $\pi \in  \{0.01,0.5\}$, we train $D$ for three iterations. For $\pi$=0.9 we found that the network would not converge for the same number of training iterations used at $\pi \in \{0.01, 0.5\}$ and so we reduced the number of iterations that $D$ was trained only once per iteration. Goodfellow et al. \cite{goodfellow2014generative} suggests that for sample generation, one iteration is sufficient.
We have demonstrated the use of $\pi \in \{0.01,0.5,0.9\}$ to show performance benefits of our proposed alternative cost function on both generative and discriminative tasks. We show that for $\pi=0.01$, a model more suitable for discriminative tasks is learned. The images generated at $\pi=0.01$ are not shown because they are either very primitive strokes or blank samples. However, at $\pi=0.1$ we find that the hypothesis still holds, whereby discriminative tasks are improved compared to regular GAN training. However, we also find that generations are both more realistic than for $\pi=0.01$ and more varied than regular GAN training. This variance comes at the cost of some samples being non-realistic. This suggests that our approach may be used to address other issues, such as lack of variation in generated samples. We leave this for future work.

\section{Conclusion}
Generative adversarial networks (GANs) are able to generate realistic looking image samples, while simultaneously learning representations for image samples from a  limited set of labelled training data. GANs are able to achieve this by minimising an adversarial cost function, which under certain conditions can be shown to approximate the Jensen-Shannon entropy. However, adversarial training can be improved, particularly when a model is intended specifically for the task of generation or classification.

We propose an alternative adversarial cost function parametrised by $\pi$ which we show to be approximately proportional to $KL[P\|Q]$, for small $\pi$ and approximately proportional to $KL[Q\|P]$ for large $\pi$. We perform both generative and discriminative tasks using our alternative cost function to show experimental evidence to support the theory motivating our alternative cost function.

Retrieval and one-shot learning experiments compared regular GAN training to training using $\pi=0.01$. Our results showed that GANs trained using our alternative cost function learned a representation for retrieval and one-shot learning that outperformed regularly trained GANs in all experiments. We also presented the first alphabet wise one-shot classification scores on the Omniglot dataset, classifying all characters in each alphabet. Previous work had only attempted to classify $5$ randomly chosen samples \cite{vinyals2016matching,santoro2016one}.

Experiments on image generation compared regular GAN training to training using $\pi=0.9$. Evidence for improved synthesis is shown by interpolating between two random samples, showing that when a GAN is trained using our alternative cost function with a large $\pi$ value, there are less gaps in the interpolation. This suggests that using our alternative cost function with a larger $\pi$ value learns a model more suitable for generation that regularly trained GANs.

Both theory and experimental results suggest that our alternative cost function, parametrised by $\pi$, allows for tuning of generative models for either generative or discriminative tasks by choosing a $\pi$ suitable for the task.


\section*{Acknowledgment}
We like to acknowledge the Engineering and Physical Sciences Research Council for funding through a Doctoral Training studentship.

\ifCLASSOPTIONcaptionsoff
  \newpage
\fi



%

\printbibliography



%





\end{document}